\pdfoutput=1
\documentclass[11pt]{article}

\usepackage[final]{coling}

\usepackage{times}
\usepackage{latexsym}

\usepackage[T1]{fontenc}
\usepackage[utf8]{inputenc}

\usepackage{microtype}

\usepackage{inconsolata}

\usepackage{graphicx}

\usepackage{amssymb}
\usepackage{xcolor}
\usepackage{booktabs}
\usepackage{amsthm,amsmath,amssymb}
\usepackage{mathrsfs}
\usepackage{tabularx}
\usepackage{multirow}
\usepackage{makecell}
\usepackage{colortbl}
\usepackage{enumitem}
\usepackage{soul} 
\sethlcolor{gray!15} % 设置高亮颜色

\title{LaERC-S:  Improving LLM-based Emotion Recognition in Conversation with Speaker Characteristics}

\author{
 \textbf{Yumeng Fu\textsuperscript{1}},
 \textbf{Junjie Wu\textsuperscript{2}},
 \textbf{Zhongjie Wang\textsuperscript{1}},
\\
 \textbf{Meishan Zhang\textsuperscript{3}},
 \textbf{Lili Shan\textsuperscript{1}},
 \textbf{Yulin Wu\textsuperscript{3}},
 \textbf{Bingquan Liu\textsuperscript{1}\thanks{Corresponding author}},
\\
 \textsuperscript{1}School of Computer Science and Technology, Harbin Institute of Technology, Harbin, China,
 \\
 \textsuperscript{2}School of Computer Science and Technology,
Soochow University, Suzhou, China,
 \\
 \textsuperscript{3}School of Computer Science and Technology, Harbin Institute of Technology, Shenzhen, China,
\\
     \href{mailto:email@domain}{24b303004@stu.hit.edu.cn, 20224027010@stu.suda.edu.cn, zjwang@insun.hit.edu.cn, }
     \\
     \href{mailto:email@domain}
     {
     mason.zms@gmail.com, 
     \{shanlili, liubq\}@hit.edu.cn, yulinwu@cs.hitsz.edu.cn}
}

\begin{document}
\maketitle
\begin{abstract}
Emotion recognition in conversation (ERC), the task of discerning human emotions for each utterance within a conversation, has garnered significant attention in human-computer interaction systems. Previous ERC studies focus on speaker-specific information that predominantly stems from relationships among utterances, which lacks sufficient information around conversations. Recent research in ERC has sought to exploit pre-trained large language models (LLMs) with speaker modelling to comprehend emotional states. Although these methods have achieved encouraging results, the extracted speaker-specific information struggles to indicate emotional dynamics. In this paper, motivated by the fact that speaker characteristics play a crucial role and LLMs have rich world knowledge, we present LaERC-S, a novel framework that stimulates LLMs to explore speaker characteristics involving the mental state and behavior of interlocutors, for accurate emotion predictions. To endow LLMs with this knowledge information, we adopt the two-stage learning to make the models reason speaker characteristics and track the emotion of the speaker in complex conversation scenarios. Extensive experiments on three benchmark datasets demonstrate the superiority of LaERC-S, reaching the new state-of-the-art.\footnote{https://github.com/bigcat-1/LaERC-S}
% \footnote{We will publicly release our dataset and source codes. \\{*}Corresponding author }
\end{abstract}

\section{Introduction}
Emotion recognition in conversation (ERC) is a fundamental task in the community of natural language processing (NLP), which targets to automatically identify the emotion of each utterance within a conversation. With the proliferation of conversation data on social media platforms, likewise Twitter and Facebook, detecting human emotions around conversations \citep{tu2022context, gao2024cept} holds promising potential for a series of real-world applications, such as recommendation \cite{song2024cagk} and opinion mining \cite{kumar2023explaining}. However, unlike sentence-level emotion recognition \cite{deng2023llms, zhang2024sentiment}, conversation involves a process of dynamic interactions, which poses a unique challenge for ERC.

\begin{figure}[t]
  \includegraphics[width=\linewidth]{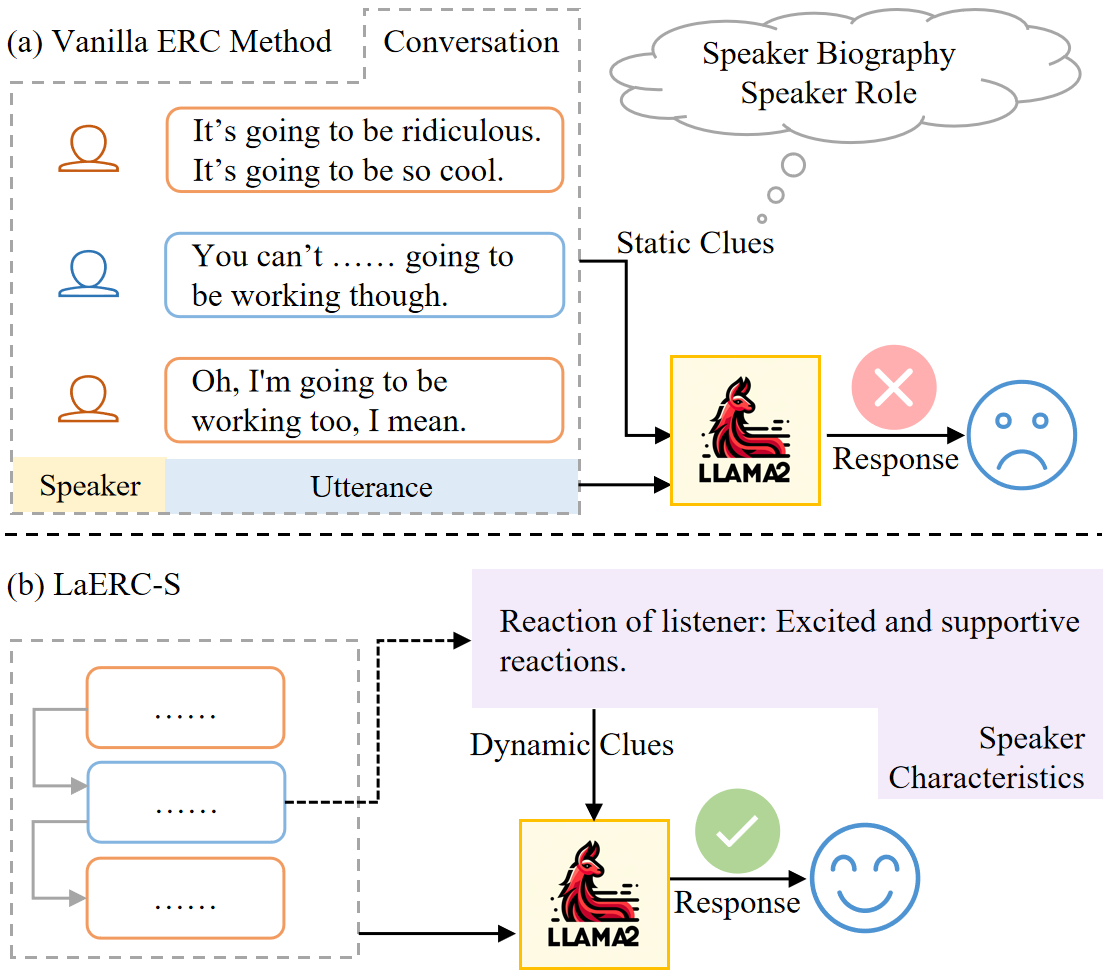}
  \caption{Comparison between existing ERC models and the proposed LaERC-S. (a) The existing ERC methods exploit static clues, such as speaker biography and speaker role, for emotional states. (b) The proposed LaERC-S captures rich and deep clues of emotional dynamics, including the mental state and behavior of interlocutors, to trigger the target emotion.}
  \label{fig:1}
\end{figure}

Faced with such a challenge, initial attempts to analyze the content of conversation relied on converstaional context modelling \cite{sun2021discourse, shen2021directed}, while current sophisticated methods \cite{song2022emotionflow, lee2022compm, zhang2023dualgats, wang2024emotion} start the investigation of speaker-specific information to mitigate emotion ambiguity. However, these methods rely on highly structured paradigms, which make the models overfit to specific data distributions, thereby hampering progress in the realm of ERC.

Apart from above studies, another strand of research resorts to the reasoning and generation capabilities of large language models (LLMs), such as PaLM \cite{chowdhery2023palm} and LLaMA2 \cite{touvron2023llama}, for different conversational datasets. A pioneering work by InstructERC \cite{lei2023instructerc} fine-tunes LLaMA2 by introducing speaker identification. Such paradigm gets significant performance compared to conventional pre-trained language models (PLMs) in ERC. Subsequently, BiosERC \cite{xue2024bioserc} integrates the biographical information of speakers to intensify LLMs-based ERC systems. As a result, the exploration of speaker characteristics can bring superior performance to their respective models.

Despite the striking results acquired by above works, they are limited by the following dilemmas: (1) Speaker identification can not provide sufficient information. (2) Speaker biography lacks clues of emotional dynamics in complex conversations. These static information makes the models tend to generate biased responses for all the utterances uttered by a certain speaker. However, as reported in \cite{hwang2021comet, Zhao2022CauAINCA}, speaker characteristics including mental state and behavior of interlocutors can provide deep and rich clues of emotional dynamics\cite{ghosal2020cosmic}, thereby triggering the target emotion. Thus, it would be beneficial to exploit such speaker characteristics into LLMs for ERC.

In this paper, we propose LaERC-S, a novel framework devised to exploit large language models and speaker characteristics for the ERC task. Specifically, we design an efficient instruction template to promote LLMs to generate the mental state, behavior and persona of interlocutors around conversations. Afterwards, to supplement LLMs with this knowledge information, we perform two-stage learning, including speaker characteristic injection and emotion recognition, for the final result. A schematic of LaERC-S is depicted in Figure~\ref{fig:1}.

Without bells and whistles, the proposed LaERC-S surpasses all ERC methods on three benchmark datasets, including IEMOCAP \cite{busso2008iemocap}, MELD \cite{poria2019meld}, and EmoryNLP \cite{zahiri2018emotion}. Moreover, LaERC-S provides a unique perspective to capture speaker characteristics in the realm of LLMs-based ERC, which can be reproduced by a sinlge GPU.

In summary, our contributions are three-fold:
\begin{itemize}
  \item We propose a simple and effective framework, namely LaERC-S, which explores large language models and speaker characteristics for emotion recognition in conversation.
  \item We design an efficient instruction template to promote LLMs to generate speaker characteristics, and adopt a two-stage learning for capturing emotional dynamics and judging emotional states in conversations.
  \item Experiments are conducted on three public datasets, including IEMOCAP, MELD, and EmoryNLP, which validates the superiority of LaERC-S over the state-of-the-art methods.
\end{itemize}

\section{Related Work}
% In this section, we successively provide the investigations of emotion recognition in conversation and speaker characteristics.

\subsection{Emotion Recognition in Conversation}
As an indispensable part of human-interaction systems, the nature of emotion recognition in conversation (ERC) refers to make the models comprehend emotion states of interlocutors within conversations, thereby generating empathy and empathic responses \cite{majumder2020mime}. In the literature, existing ERC studies \cite{poria2017context, majumder2019dialoguernn, ghosal2019dialoguegcn, li2021past, li2023skier, Zhao2022CauAINCA, zhang2023dualgats, tu2023empirical, jian2024emotrans} can be roughly divided into two ideas. One relys on pre-trained language models (PLMs) to model conversational context and speaker for emotion prediction. Typically, DialogXL \cite{shen2021dialogxl} introduces an enhanced memory to store conversational contexts, and further captures intra- and inter-speaker dependencies for multi-party structures. CEPT \cite{gao2024cept} devises a mixed prompt template and a label mapping strategy for conversational contexts and comprehensive emotions, respectively. With the advancements of pre-trained large language models (LLMs), another line of research attempts to employ LLMs to the task of ERC. Recently, InstructERC \cite{lei2023instructerc} transforms ERC into a retrieved-based Seq2Seq form for LLMs adaptation. BiosERC \cite{xue2024bioserc} leverages speakers' personalities to enhance LLMs.

These methods reveal the statement that speaker characteristics are beneficial for emotion recognition in conversation. However, they lack convincing interpretations for acquiring speaker-specific information, thereby limiting emotional expressions. Therefore, in this paper, we attempt to adopt an explainable way to explore large language models and speaker characteristics for the ERC task.

\subsection{Speaker Characteristics}
Speaker characteristics involve the mental state, behavior and persona of interlocutors in social interaction \cite{bosselut2019comet, sap2019atomic, hwang2021comet}. It is beneficial for a human-computer interaction system to comprehend the speaker's intention and purpose, as well as analyze situationally-relevant speaker's reaction and behavior. Motivated by such superiority, a series of works employ speaker characteristics to numerous downstream tasks, such as question answering \cite{zhang2022greaselm}, empathic response generation \cite{sabour2022cem}, and emotional gold mining \cite{wang2024ecok}. In recent years, scholars have paid attention to making progress in ERC by exploring speaker characteristics. These studies leverage conversational relations expressed by a triplet form, to learn the interaction between speakers. Typically, COSMIC \cite{ghosal2020cosmic} exploits a commonsense knowledge base to learn commonsense features for emotion prediction. SKAIG \cite{li2021past} constructs a graph to capture speaker's psychological states. CauAIN \cite{Zhao2022CauAINCA} regards commonsense knowledge as causal clues to trigger the target emotion.

Our method is different from these methods that achieve speaker characteristics from relationships among utterances. In this paper, we extract rich world knowledge from LLMs by devising an efficient template while making the models reason speaker characteristics and track emotional states. This stimulates the proposed LaERC-S to provide more accurate emotion predictions.

\begin{figure*}[t]
  \includegraphics[width=\linewidth]{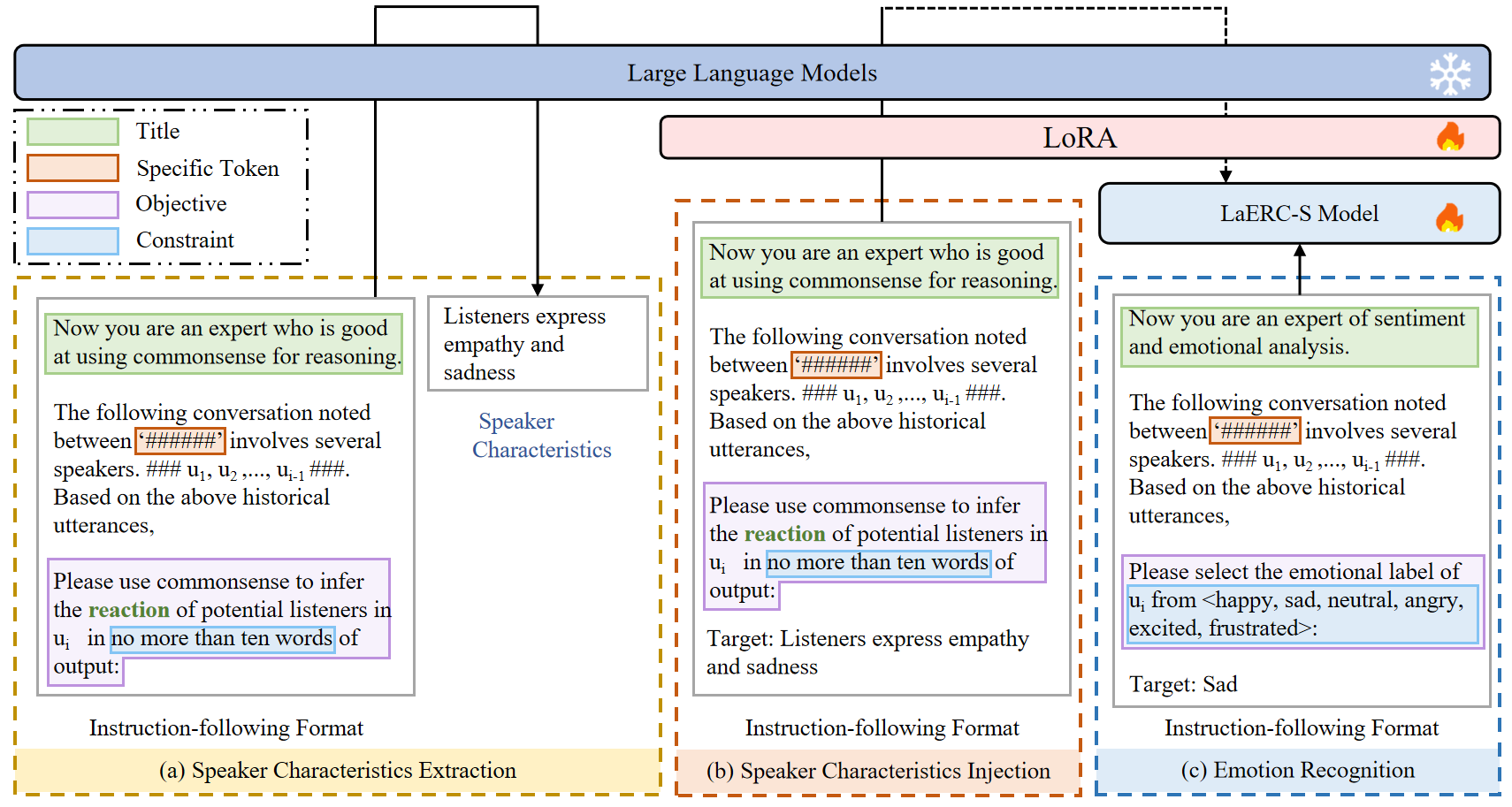}
  \caption {The overview of LaERC-S. LaERC-S includes speaker characteristics extraction and injection, emotion recognition. In the speaker characteristics extraction, speaker characteristics are  extracted from LLMs. In the speaker characteristics injection, the generated speaker-characteristics are employed to make the models perceive emotional dynamics. In the emotion analysis, the conversational contents and predefined emotional labels are converted into a formatted input for the final response. As depicted in the instance, LaERC-S bridges the gap between speaker characteristics and the response of ``sad''.}
  \label{fig:2}
\end{figure*}

\section{The LaERC-S Framework}
In this section, we present a framework, namely LaERC-S, which introduces speaker characteristics for adapting LLMs to emotion recognition in conversation, as shown in Figure~\ref{fig:2}. First, we provide the vanilla model in the task of ERC, followed by the specifics of LaERC-S, including speaker characteristic extraction and injection, emotion recognition. Moreover, LaERC-S can also be extended to any of mainstream large language models.

\subsection{Vanilla ERC Model}
A conversation data source as $\mathcal{D}=\left\{(C_i, Y_i)\right\}_{i=1}^N$, where the symbol $C_i$ denotes the $i$-th conversation, and $N$ is the size of $\mathcal{D}$. Each conversation includes a sequence of utterances $\mathcal{U}=\left\{u_j\right\}_{j=1}^S$, where the sign $S$ is the number of all utterances. Each utterance in a conversation is assigned with a ground truth label $y_j\in\left\{e_1,e_2,...,e_K\right\}$, where $K$ is the number of emotion categories.

Generally, the ERC model $\mathcal{M}$ based on LLMs is learned from $\mathcal{D}$ to provide a response $r$ over a set of the predefined emotion labels $\mathcal{E}=\left\{e_k\right\}_{k=1}^K$. The whole process can be expressed as follows:
\begin{eqnarray}\label{eq:vcg_o}
  r_{j,i} = \mathcal{M}(u_{<j}, u_j, \mathcal{E}),
\end{eqnarray}
\noindent where, $u_{<j}$ denotes the historical utterances before the target utterance $u_j$ in the $i$-th conversation.

\subsection{Speaker Characteristic Extraction}

To extract high quality speaker characteristics in conversation, we adopt prompt engineering for extraction due to the beneficial of this technology \cite{liu2023pre, white2023prompt,giray2023prompt}. Besides, considering the fact that pre-trained LLMs serve as a rich world knowledge base, we design a template to query LLMs to capture speaker characteristics. Besides, we manually verified speaker characteristics extracted from the large model. We provide the generation procedure of available information regarding speaker characteristics in conversational scenarios, as depicted in Figure~\ref{fig:2} (a).

Typically, we investigate previous studies \cite{sap2019atomic, hwang2021comet}, and observe that speaker characteristics cover mental state, behavior and persona. Appendix~\ref{A.3.1} presents the definitions of the information. Mental state reflects emotional states of interlocutors, containing three relations, i.e., `oReact', `xReact' and `xIntent'. Behavior means a response to an event, including `xNeed', `xWant', `oWant', `xEffect' and `oEffect'. Persona indicates the interlocutor's attribute by `xAttr'.

These key elements from different perspectives reveal the interaction between utterances, which is intuitively projected into the query template for retrieving available information regarding speaker characteristics. The templates relevant to all the key elements are presented in Appendix~\ref{A.3.2}.

% Before presenting LaERC-S, we provide the generation procedure of available information regarding speaker characteristics in conversational scenarios, as depicted in Figure~\ref{fig:2} (a). Consider the fact that pre-trained LLMs serve as a rich world knowledge base, we design a template to query LLMs to capture speaker characteristics for ERC.

% Typically, we investigate previous studies \cite{sap2019atomic, hwang2021comet}, and observe that speaker characteristics cover mental state, behavior and persona. Mental state reflects emotional states of interlocutors, containing three relations, i.e., `oReact', `xReact' and `xIntent'. Behavior means a response to an event, including `xNeed', `xWant', `oWant', `xEffect' and `oEffect'. Persona indicates the interlocutor's attribute by `xAttr'.

% These key elements from different perspectives reveal the interaction between utterances, which is intuitively projected into the query template for retrieving available information regarding speaker characteristics. The templates relevant to all the key elements are presented in Appendix~\ref{sec:appendix_1}.

\subsection{Speaker Characteristic Injection}
Speaker characteristic injection is to learn clues of emotional dynamics in conversation scenarios, which endows the model with speaker characteristics for subsequent emotion analysis. Although pre-trained large language models cover speaker-specific information, they have not yet been activated the perception capability about this under conversational contexts. To this end, we adopt a instruction-tuning strategy tailored to endow LLMs with speaker characteristics at the initial stage, as shown in Figure~\ref{fig:2} (b).

Typically, we design an instruction template with a certain key element and basic elements for knowledge analysis. A key element is one of any relationships provided by above preliminary. The basic elements comprises four aspects, i.e., `title', `specific token' and `objective', `constraint'. The `title' indicates that the role of LLMs expert apt in learning emotional clues in conversations. The `specific token' is to separate conversation contents. The `objective' refers to a concise elucidation of the task of knowledge analysis, which provides a response based on conversation contexts. The `constraint' is used to limit the length of the response for avoiding hallucinations. For reference, we construct the input template to align with the instruction-following template of information retrieval at preliminary.
% \begin{quote}
%   \textit{``Now you are an expert who is good at using commonsense for reasoning. The following conversation noted between `\#\#\#\ \#\#\#' involves several speakers. \#\#\# $u_1$, $u_2$, ... , $u_{i-1}$ \#\#\#. Based on the above historical utterances. Please use commonsense to infer the {[oReact]} of potential listeners in $u_i$, no more than ten words of output: ''}
% \end{quote}

\subsection{Emotion Recognition}
After the above stage, we achieve an initial model that is available to perceive clues of emotional dynamics in conversations. However, there is a gap between these clues and emotion states. To reach this, we further conduct an instruction-tuning strategy to learn the interplay between emotional tendencies and clues, as depicted in Figure~\ref{fig:2} (c).

To aligned with the initial stage, we make adjustments in the initial instruction-following template, i.e., title, objective and constraint. Typically, the ``title'' presents the role of LLMs as assistant skilled in sentiment and emotion analysis. The ``objective'' proposes to give a emotional label for the target utterance in a conversation. The ``constraint'' refers to a set of the predefined emotional labels. Such format can maximize the mutual synergy between multiple tasks, while the generated knowledge information does not need to be added into this template without additional computing resources.

Overall, the objective function for various tasks can be defined as follows:
\begin{eqnarray}\label{eq:vcg_1}
  L_{k} = \sum_{i'}^{j} - \mathrm{log}P(\mu_{(k,i')}|x_k,\theta_{k}),
\end{eqnarray}
\noindent where $k$ indicates a certain stage, and $x_k$ is the instruction-following template to the certain stage. $\mu_{(k,i')}$ denotes the generated token. In addition, $\theta_{k}$ denotes the trainable parameters in LLMs.

Finally, after the second stage, the well-trained model is leveraged for inference purposes. We choose `oReact` item as the final LaERC-S model for emotion analysis in conversation.
% Please refer to see Appendix~\ref{sec:appendix_2} for instruction-following templates in the process of each stage.

\section{Experiments}
In this section, we successively present three commonly used conversation datasets, compared baselines and basic experimental settings, and then analyze the experimental results in detail.

\subsection{Datasets}
% To evaluate LaERC-S for the task of ERC, we adopt three public datasets, including IEMOCAP \cite{busso2008iemocap}, MELD \cite{poria2019meld}, and EmoryNLP \cite{zahiri2018emotion}.
We evaluate LaERC-S on three representative datasets which involve
 IEMOCAP \cite{busso2008iemocap}, MELD \cite{poria2019meld}, and EmoryNLP \cite{zahiri2018emotion}. More details about these datasets can be found in 
 Appendix~\ref{A.1}.

% \noindent \textbf{IEMOCAP} is a dataset collected from improvisations or scripted scenarios, which contains 12 hours of conversation videos from 10 unique speakers. It has five sessions consisting of 151 conversations and 7,433 utterances. Each utterance is annotated with one of six emotion classes: neutral, happy, sad, excited, angry, and frustrated. 

% \noindent \textbf{MELD} is another dataset including more than 13,000 video snippets from the Friends TV series. It comprises 1,433 conversations and 13,708 utterances in total. Each utterance is labeled as one of seven emotion categories: anger, disgust, fear, joy, neutral, sadness, and surprise. 

% \noindent \textbf{EmoryNLP} is also based on the Friends TV series, which contains 97 episodes, 897 scenes and 12,606 utterances. Each utterances is annotated as one of seven emotion types: neutral, joyful, peaceful, powerful, scared, mad, and sad.

\subsection{Baselines}
To demonstrate the superiority of LaERC-S in the task of emotion recognition in conversation, we compare LaERC-S with two kinds of mainstream ERC methods as follows.

\noindent (i) Conventional ERC methods: COSMIC \cite{ghosal2020cosmic}, SKAIG \cite{li2021past}, DialogXL \cite{shen2021dialogxl}, SPCL \cite{song2022supervised}, CauAIN \cite{Zhao2022CauAINCA},  DualGATs \cite{zhang2023dualgats}, MKFM \cite{tu2023empirical}, MFAM\cite{hou2023enhancing}, and CEPT \cite{gao2024cept}.

\noindent (ii) LLMs-based ERC methods: ChatGPT \cite{ouyang2022training}, InstructERC \cite{lei2023instructerc}, and BiosERC \cite{xue2024bioserc}.

% Besides, for evaluation, we follow previous ERC methods, using the weighted-F1 score to measure the performance of our proposed method and all the listed methods.

\subsection{Implementation Details}
Following current LLMs-based ERC methods \cite{lei2023instructerc, xue2024bioserc}, we adopt the LLaMA2 \cite{touvron2023llama} as the foundational model in this paper. Consider the expensive training costs and the issue of catastrophic forgeting, we use a lightweight training technique, i.e., LoRA \cite{hulora}, to stay the model weights frozen and train a small portion of model parameters for specific subtasks. In detail, we set the learning rate to 2e-4, and the converstaional context window to 12 for all evaluation datasets. In the first stage, the batch size is set to 8. In the second stage, the batch size is set to 16. For the hyper-parameter such as epoch, we tune them on the development dataset. The reported results are an average over five random runs. All the experiments are implemented by using PyTorch \cite{paszke2019pytorch} on a single NVIDIA Tesla V100 GPUs. We restrict the input length to 1024. More details about parameters analysis of context window can be found in Appendix~\ref{window}. 

\begin{table}
  \centering
  \setlength{\tabcolsep}{0.9pt}
  \begin{tabular}{l|cccc}
    \hline
    Methods           & IEMOCAP & MELD & EmoryNLP & Avg.  \\
    \hline
    COSMIC & 63.43 & 65.03 & 38.49 & 55.65 \\
    SKAIG & 66.96 & 65.18 & 38.88 & 57.01 \\
    DialogXL & 66.20 & 62.41 & 34.73 & 54.45 \\
    SPCL & 69.21 & 66.13 & 40.25 & 58.53 \\
    CauAIN & 65.01 & 64.89 & 37.87 & 55.92 \\
    DualGATs & 67.68 & 66.90 & 40.29 & 58.29 \\
    MKFM & 68.08 & 65.50 & 39.76 & 57.78 \\
    MFAM & 70.16 & 66.65 & 41.06 & 59.29 \\
    CEPT & 70.53 & 67.51 & - & - \\
    \hline
    ChatGPT & 40.07 & 54.37 & 37.55 & 44.00 \\
    BiosERC & 69.02 & 68.72 & 41.44 & 59.73 \\
    InstructERC & 71.39 & 69.15 & 41.37 & 60.64 \\
    \hline
    LaERC & 69.95 & 68.86 & 40.87 & 59.89 \\
    \rowcolor{gray!15} LaERC-S & \textbf{72.40} & \textbf{69.27} & \textbf{42.08} & \textbf{61.25} \\
    \hline
  \end{tabular}
  \caption{\label{tab:1}
  Performance comparison between our proposed LaERC-S and existing ERC methods on three conversation datasets. LaERC is finetuning Llama2-7B to recognize emotion in conversation. The p-values are all below 0.001 by using pairwised t-test towards our method and the corresponding baselines. The best results are \textbf{bolded}. Our evaluation metric is weighted-F1. 
  }
\end{table}

\subsection{Main Results}
To illustrate the effectiveness of LaERC-S framework in the task of ERC, we report the performance of our proposed method and other baseline methods in Table~\ref{tab:1}, where `Avg.' denote the overall average performance on three benchmark datasets. We can observe that our proposed LaERC-S achieves the best results than other all methods on three public datasets. Such performance demonstrates that LaERC-S has stronger generalization and more accurate predictions for emotion recognition.

Typically, compared to previous ERC paradigms, LLMs-based ERC methods have achieved significant results than them. The reason is the thorough understanding capability of pre-trained large language models. Notably, our proposed method LaERC-S achieves an improvement of 1.01\% over InstructERC, 3.38\% over BiosERC on the IEMOCAP dataset, respectively. For more complex conversation scenarios, such as MELD and EmoryNLP datasets, LaERC-S still provides meaningful gains in performance. This is due to the efficiency of speaker characteristics explored from the key element `oReact' in the proposed LaERC-S.

Besides, we notice that the results of ChatGPT in zero-shot scenarios are far from other methods that trained with the full dataset. It is attributed to the purpose of universality rather than specific tasks. Therefore, consistent with LaERC-S, it is essential to fine-tune the models for the task of ERC. In summary, the above comparative results present that LaERC-S outperforms all the ERC methods.

\subsection{Ablation Study}

In this section, we demonstrate the superiority of the proposed method LaERC-S from the impact of speaker characteristics. It is to measure the importance of introducing speaker characteristics, and how to sufficiently exploit it in the task of ERC. The experimental results are presented in Table~\ref{tab:2}, we can achieve the following findings:

% In this section, we demonstrate the superiority of the proposed method LaERC-S from two perspectives, including the impact of speaker characteristics and key elements. The former is to measure the importance of introducing speaker characteristics, and how to sufficiently exploit it in the task of ERC. The latter targets to examine the effectiveness of various key elements in LaERC-S. The experimental results are presented in Table~\ref{tab:2} and Table~\ref{tab:3}, we can achieve the following findings:

\begin{table}[t]
  \centering
  \setlength{\tabcolsep}{1.3pt}
  \begin{tabular}{l|cccc}
    \hline
    Methods & IEMOCAP & MELD & EmoryNLP & Avg. \\
    \hline
    w/o S & 69.95 & 68.86 & 40.87 & 59.89 \\
    w M & 70.21 & 68.52 & 41.49 & 60.07 \\
    \hline
    w R & 71.43 & 69.04 & 40.82 & 60.43 \\
    \rowcolor{gray!15} w S & 72.40 & 69.27 & 42.08 & 61.25 \\
    \hline
  \end{tabular}
  \caption{\label{tab:2}
  Performance comparison by speaker characteristics in emotion recognition. `M' refers to directly introduce the generated speaker characteristics into the stage of emotion analysis. `R' and `S' regard speaker identification and speaker characteristics injection as the initial stage.}
\end{table}

% \begin{table}[t]
%   \centering
%   \setlength{\tabcolsep}{1.3pt}
%   \begin{tabular}{l|cccc}
%     \hline
%     Methods & IEMOCAP & MELD & EmoryNLP & Avg. \\
%     \hline
%     % w/o S & 69.95 & 68.86 & 40.87 & 59.89 \\
%     % w M & 70.21 & 68.52 & 41.49 & 60.07 \\
%     % \hline
%     w L & 69.27 & 69.04 & 41.21 & 60.43 \\
%     \rowcolor{gray!15} w S & 72.40 & 69.27 & 42.08 & 61.25 \\
%     \hline
%   \end{tabular}
%   \caption{\label{tab:2}
%   Performance comparison by speaker characteristics in emotion recognition. `M' refers to directly introduce the generated information into the stage of emotion analysis. `R' and `S' regard speaker identification and knowledge analysis as the initial stage.}
% \end{table}

\begin{itemize}[itemsep=2pt,topsep=2pt,parsep=0pt,leftmargin=*]
  \item To understand the importance of introducing information around speaker characteristics in conversational scenarios, we present the results of relevant experiments in Table~\ref{tab:2}, where the first two rows are the one-stage learning, and the last two rows are the two-stage learning. For reference, in the first row of this table, we eliminate any of speaker characteristics, and solely implement the stage of emotion analysis, presenting a lowest result. Next, we directly incorporate the generated speaker characteristics into the stage of emotion analysis, resulting in the performance improvements in the most of datasets. This highlights the importance of speaker characteristics in ERC.
  \item On the other hand, we adopt the two-stage learning strategy, and regard speaker identification as the initial stage before the stage of emotion analysis. Such method outperforms the first two methods (i.e., one-stage learning), suggesting the efficiency of two-stage learning in the ERC task. In the last row of Table~\ref{tab:2}, we present the final performance of LaERC-S, which achieves the best results on all the datasets. These experiments demonstrate that LaERC-S can achieve accurate emotion predictions through introducing speaker characteristics, and use the two-stage learning to magnify the efficiency of speaker characteristics to enhance the model in performance.

  % \item Further, to investigate the influence of different key elements of the initial stage in the proposed method LaERC-S, we present a serise of fine-grained experiments in Table~\ref{tab:3}, where `Key Ele.' denotes the key elements. From this table, apart from the key element `xAttr', the remaining key elements can efficiently bring performance improvements compared to LaERC-S without the initial stage (i.e., the first row of Table~\ref{tab:2}). This is attributed to the fact that the key element `xAttr' reflects the attribute of a certain person under a given event, which struggles to capture clues of emotional dynamics in conversation scenarios. In contrast, the extracted information from the perspectives of mental state and behavior can provide rich and deep clues of emotional dynamics to the model for emotion prediction, which is consistent with previous ERC studies. In the type of mental state, `oReact' describes the reaction of listener that refers to the interlocutor of the target utterance in a conversation. This rely on the conversation context provides dynamic clues regarding emotional state, leading to a significant improvement in performance. Therefore, in this paper, we choose `oReact' as the key element in the initial stage of the proposed LaERC-S method for emotion recognition in conversation.
\end{itemize}

\begin{table}[t]
  \centering
  \setlength{\tabcolsep}{1.8pt}
  \begin{tabular}{l|cccc}
    \hline
    Key Ele. & IEMOCAP & MELD & EmoryNLP & Avg. \\
    \hline
    \rowcolor{gray!15} oReact  & 72.40 & 69.27 & 42.08 & 61.25 \\
    xIntent & 71.60 & 69.56 & 41.39 & 60.85 \\
    xReact  & 71.14 & 69.17 & 39.91 & 60.07 \\
    \hline
    xEffect & 70.70 & 68.54 & 41.94 & 60.39 \\
    oEffect & 71.27 & 68.27 & 41.64 & 60.39 \\
    oWant   & 70.81 & 68.87 & 43.24 & 60.97 \\
    xWant   & 71.24 & 68.65 & 42.37 & 60.75 \\
    xNeed   & 71.94 & 68.50 & 40.27 & 60.24 \\
    \hline
    xAttr   & 70.08 & 67.82 & 40.54 & 59.48 \\
    \hline
  \end{tabular}
  \caption{\label{tab:3}
  Analysis of different elements in the initial stage of LaERC-S. `oReact' is regarded as the final LaERC-S model for emotion analysis in conversation.}
\end{table}

\section{In-depth Analysis}

\subsection{Elements Selection}
% Elements selection targets to examine the effectiveness of various key elements in LaERC-S. 
% The experimental results are presented in  Table~\ref{tab:3}, we can achieve the following findings:
To investigate the influence of different key elements (Key Ele. for short) within the speaker characteristics extraction and injection stage, we design a more detailed experiment by leveraging just one key elements. 
\par
Table~\ref{tab:3} shows the results, from which we can observe that
 apart from `xAttr', others can efficiently bring performance improvements compared to LaERC-S without the initial stage (the first row of Table~\ref{tab:2}). 
 These phenomena can be attributed to the fact that `xAttr' only reflects the personal attribute, which struggles to capture dynamic emotional clues in conversation scenarios. 
 And conversely, the extracted information from the mental state and behavior can provide richer and deeper dynamic emotional clues for emotion prediction~\citep{li2021past, ghosal2020cosmic}. 
 \par
 Notably, in mental state, `oReact' describes the reaction of listener that refers to the interlocutor of the target utterance in a conversation. It is manifested as dynamic emotional clues provided by the conversational context, capable of revealing emotional states, leading to 
a significant improvement in performance.
 % This key element will generate a dynamic emotional clues, which can lead a a significant improvement on perfomance.
 %
 % %
 % As the most relevant key element to the ERC task, it leads a significant improvement in performance.
 %
 Therefore, we choose `oReact' as the key element in the initial stage.

% \subsection{Parametric Sensitivity Analysis}

%---------------------------
%存在的问题：没有下降的点，(1) InstructERC只有20的结果，没有12-20的中间值，但是我们开不到这么大。
% \begin{table*}[t]
%   \centering
%   \setlength{\tabcolsep}{5.3pt}
%   \begin{tabularx}{\linewidth}{c|cc|cc|cc}
%   \hline
%   \multirow{2}{*}{\makecell{Context\\Window}} & \multicolumn{2}{c|}{IEMOCAP} & \multicolumn{2}{c|}{MELD} & \multicolumn{2}{c}{EmoryNLP}\\
%   \cline{2-7}
%   & InstructERC  & LaERC-S  & InstructERC  & LaERC-S & InstructERC  & LaERC-S \\
%   \hline
%   1    & 56.12 &	61.13 (5.01$\uparrow$)  &	65.91 &	68.70 (2.79$\uparrow$) &	38.32 &	39.84 (1.52$\uparrow$)  \\
%   5   & 68.65 &	69.97 (1.32$\uparrow$) &	66.97 &	69.21 (2.24$\uparrow$) &	40.48 &	41.96 (1.48$\uparrow$) \\
%   12    & 71.39 &	72.40 (1.01$\uparrow$) &	69.15 &	69.27 (0.12$\uparrow$) &	41.37 &	42.08 (0.71$\uparrow$)  \\
%   \hline
%   \end{tabularx}
%   \caption{Parameter analysis of the context window in the proposed method LaERC-S on three widely-used benchmark datasets. The symbol $\uparrow$ represents an improvement in performance over the compared method InstructERC.}
%   \label{tab:4}
% \end{table*}

\begin{table}[t]
  \centering
  \setlength{\tabcolsep}{.8pt}
  \begin{tabularx}{\linewidth}{l|cccc}
    \hline
    Models & IEMOCAP & MELD & EmoryNLP & Avg. \\
    \hline
    Baseline & 69.95 & 68.86 & 40.87 & 59.89 \\
    \hline
    Mistral-7B     & 70.44 & 69.15 & 41.25 & 60.28 \\
    Mixtral-7B     & 70.86 & 69.32 & 40.88 & 60.35 \\
    Claude      & 70.88 & 69.22 & 41.77 & 60.62 \\
    \hline
    % Llama3-70B  & 70.58 & 68.57 & 39.84 & 59.66 \\
    Llama2-13B  & 70.31 & 69.58 & 43.19 & 61.03 \\
    \rowcolor{gray!15} Llama2-7B   & 72.40 & 69.27 & 42.08 & 61.25 \\
    \hline
  \end{tabularx}
  \caption{\label{tab:5}
  Performance of LaERC-S with different large language models on three public conversation datasets. `Claude' represents Claude-3-Haiku. }
\end{table}

% \newpage
%重新设置小章节，然后把每个表格的分析放入，分析里面的细节还需要进一步微调，待修改
% \subsection{The Context Window Investigation}
% To examine the impact of the context window in the performance, we conduct a parametric sensitivity analysis with different context window, as depicted in Table~\ref{tab:4}. We can notice that LaERC-S achieves the superior performance over InstructERC under any context window settings. This highlights the efficiency of LaERC-S on the task of ERC. For reference, in the first row of the table, LaERC-S provides a 5.01\%, 2.79\%, and 1.52\% improvements over InstructERC on IEMOCAP, MELD, and EmoryNLP, respectively. With the size increasing of the context window, the performance of both methods presents a tendency of improvement. Compared with MELD and EmoryNLP, the models on the dataset IEMOCAP present a significant improvement with the same context window. This is attributed to the average length of conversation in various datasets. The average length of IEMOCAP is longer than that of other datasets, thereby exploiting the larger window providing the necessary historical context for an improvement in performance. Although the performance discrepancy between them gradually decreases, the proposed LaERC-S still achieves significant superiority on three benchmark datasets. Therefore, we set a context window of 12 in LaERC-S to sufficiently capture the historical cotext in a conversation.

\begin{table}[t]
  \centering
  \setlength{\tabcolsep}{1.5pt}
  \begin{tabularx}{\linewidth}{l|cccc}
  \hline
  Templates & IEMOCAP & MELD & EmoryNLP & Avg. \\
  \hline
  Template\ 1	& 71.86 &	68.32 &	40.62  & 60.27 \\
  Template\ 2 &	71.54 &	69.05  &	40.25 & 60.28\\
  Template\ 3	& 71.85 &	68.04 	&41.51 & 60.47 \\
  \rowcolor{gray!15} Template\ 4	& 72.40 & 	69.27 &	42.08 & 61.25 \\
  \hline
  \end{tabularx}
  \caption{\label{tab:6}
  Performance of LaERC-S with different templates on three benchmark datasets.}
\end{table}

% \begin{figure*}[t]
%     \centering
%   \includegraphics[width=0.8\linewidth]{case_study.jpg}
%   \caption {The case study of four samples from IEMOCAP dataset.}
%   \label{fig:3}
% \end{figure*}

% \begin{figure*}[t]
%   \includegraphics[width=0.4\linewidth]{case_study.jpg}
%   \caption {A minimal working example to demonstrate how to place
%     two images side-by-side.}
%   \label{fig:4}
% \end{figure*}

%+开源模型的参考文献？
% \section{In-depth Analysis}
\subsection{Different LLMs Impact on speaker Characteristic Extraction}
To demonstrate the expansibility of LaERC-S, we make a comprehensive comparison of the generated speaker characteristics from different large language models with parameters ranging from 7B to 13B, as shown in Table~\ref{tab:5}. Specifically, we employ a series of representative LLMs including Mistral-7B \cite{jiang2023mistral}, 
% {Mistral-7B}\footnote{\url{https://mistral.ai/}}, 
% % {Mixtral-7B}\footnote{\url{https://mistral.ai/}}, 
% % {Claude}\footnote{\url{https://claude.ai/}}, ,
% (Mistral-8\times \text{7B}), 
Mixtral-8$\times$7B 
 \cite{jiang2024mixtral},
Claude-3-Haiku,
Llama2 (13B, 7B) \citep{touvron2023llama}, for evaluation. In the first row of the table, we present the performance of the baseline to intuitively understanding the impact of speaker characteristics in LaERC-S. We can see that various LLMs generate the speaker characteristics that is beneficial to provide the performance improvements of the proposed method. This emphasize the expansibility of the LaERC-S. Moreover, we intuitively think the reason why Llama3-13B performs worse than Llama2-7b is the inconsistency of the adopted models between extraction and injection. The larger-scale language models have not yet provided significant improvements in performance. However, LaERC-S employs Llama2-7B to generate speaker characteristics and further train it for more accurate emotion predictions.

\subsection{Different Template Impact on 
Speaker Characteristics Generation}

To explore the impact of various templates in performance, we conduct experiments with four different templates (more details about the templates can be found in Appendix~\ref{sec:appendix_3} ), as presented in Table~\ref{tab:6}.
We randomly sampled 100 samples from the training set and generated speaker characteristics for each instance using four different templates. We manually validated the quality of the speaker characteristics produced for sample to determine which template to select. Among the 100 samples, we discovered that 80\% 
selected Template 4, 8\% selected Template 3, 7\% selected Template 2, and 5\% selected Template 1.

Specifically, Although each template solely exists subtle discrepancies, they present different results. For instance, the word ``potential'' in template 4 is removed in template 2, leading to a 0.97\% drop in performance, suggesting the importance of the template
in LaERC-S. These experiments proves that LLMs are sensitive to templates, which validates that a good template is important in LaERC-S for emotion recognition task. Therefore, we choose the template 4 in LaERC-S to perform more accurate emotion predictions.

\begin{figure*}[t]
    \centering
  \includegraphics[width=0.9\linewidth]{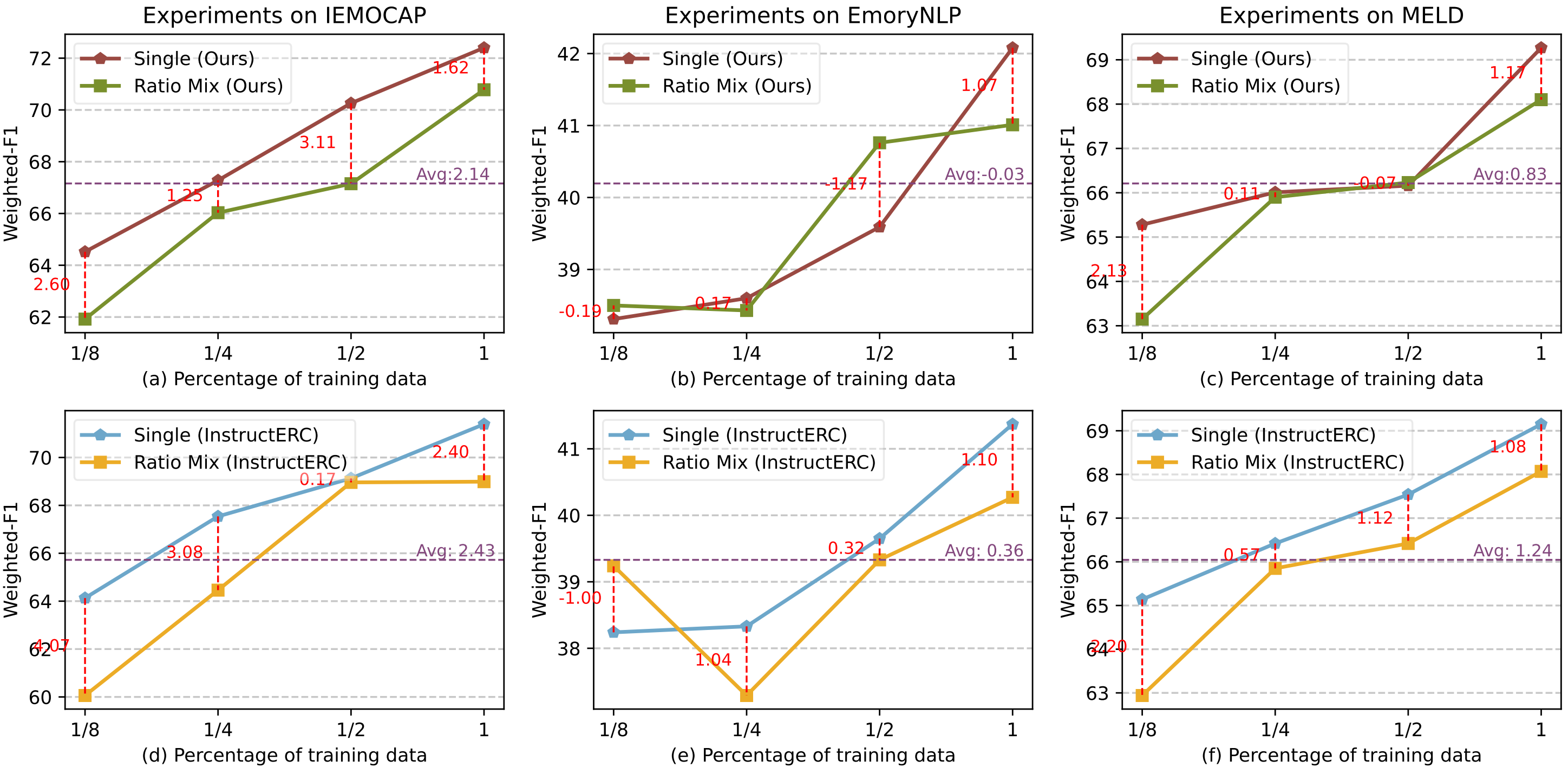}
  \caption {The cross-datasets analysis. 
  % We sequentially select data from each dataset in the ratios of 1/8, 1/4, 1/2, and 1. 
  % and 1. Avg represents the average of difference of single and ratio mix under different data configurations. 
  % `Single' means we train on single dataset. `Ratio Mix' means we train on ratio mix dataset.
  `Single' and `Mixed Ratio' refer to training on a single and mixed dataset, respectively.
 We sequentially select data from each dataset in the ratios of 1/8, 1/4, 1/2, and 1. 
  `Avg' represents the average of the differences between `Single' W-F1 and `Ratio mix' W-F1. }
  \label{figure_mix} 
\end{figure*}

\subsection{Robustness Analysis}
\label{RB}
\par
To validate the robustness capability of LaERC-S, we conduct a cross-dataset validation experiment. 
% Corpus generated from real-world conversations is diverse, which poses challenges for the robustness of ERC models.
% 
% To validate the generalization capability of LaERC-S, we conduct a cross-dataset validation experiment. 
% and choose InstructERC as our strong baseline due to its outstanding performance compared to other ERC models.
\par 
% Specifically, within the settings of this experiment, all emotional labels across the datasets are standardized.[ap]
Specifically, we first extract data with the same proportion from the training sets of three datasets, and then merge them into a mixed dataset. 
Subsequently, we train LaERC-S on the mixed dataset and inference on the test sets of the three original datasets. 
Finally, we demonstrate the generalization of LaERC-S by comparing its weighted-F1 score to that obtained from training and inference both on the original dataset. 
Notably, in this experiment, we choose InstructERC as our strong baseline due to its outstanding performance compared to other previous ERC models. 
% Finally, in Table~\ref{}(a)(b)(c), we find experiments on mixed datasets only have experienced a minor drop, which demonstrate the robustness of LaERC-S by comparing its weighted-F1 score to that obtained from training and inference both on the original single dataset. 

% Besides, Table~\ref{}(a)(b)(c) compared with Table~\ref{}(d)(e)(f), we find The variance between the results of LaERC-S using single and mixed datasets is less than that of InstructERC.
%
% Notably, in this experiment, we choose InstructERC as our strong baseline due to its outstanding performance compared to other previous ERC models. 
%
\par
The results are shown in Figure~\ref{figure_mix}, from which we can observe that LaERC-S is less affected by the cross-dataset validation compared to InstructERC. 
More specifically, in the dataset IEMOCAP and EmoryNLP, the `Avg' of the proposed LaERC-S surpasses the baseline method InstructERC by significant improvements of  0.29\% and 0.39\%, respectively. Even in the more complex conversation dataset MELD, LaERC-S presents a better robustness (a performance improvement of 0.41\%).
These phenomena exhibits the exceptional robustness of our model.
More details about robustness analysis  can be found in Appendix~\ref{A.4}.

% \begin{figure*}[t]
%     \centering
%   \includegraphics[width=0.8\linewidth]{case_study.jpg}
%   \caption {The case study of four samples from IEMOCAP dataset.}
%   \label{fig:3}
% \end{figure*}

% \begin{figure*}[t]
%     \centering
%   \includegraphics[width=0.9\linewidth]{mix.jpg}
%   \caption {The cross-datasets analysis. 
%   % We sequentially select data from each dataset in the ratios of 1/8, 1/4, 1/2, and 1. 
%   % and 1. Avg represents the average of difference of single and ratio mix under different data configurations. 
%   % `Single' means we train on single dataset. `Ratio Mix' means we train on ratio mix dataset.
%   `Single' and `Mixed Ratio' refer to training on a single and mixed dataset, respectively.
%  We sequentially select data from each dataset in the ratios of 1/8, 1/4, 1/2, and 1. 
%   `Avg' represents the average of the differences between `Single' W-F1 and `Ratio mix' W-F1. }
%   \label{figure_mix} 
% \end{figure*}

% \begin{figure*}[t]
%     \centering
%   \includegraphics[width=0.85\linewidth]{case_study.jpg}
%   \caption {The case study of four samples from IEMOCAP dataset.}
%   \label{fig:3}
% \end{figure*}

\begin{figure}[t]
    \centering
  \includegraphics[width=0.85\linewidth]{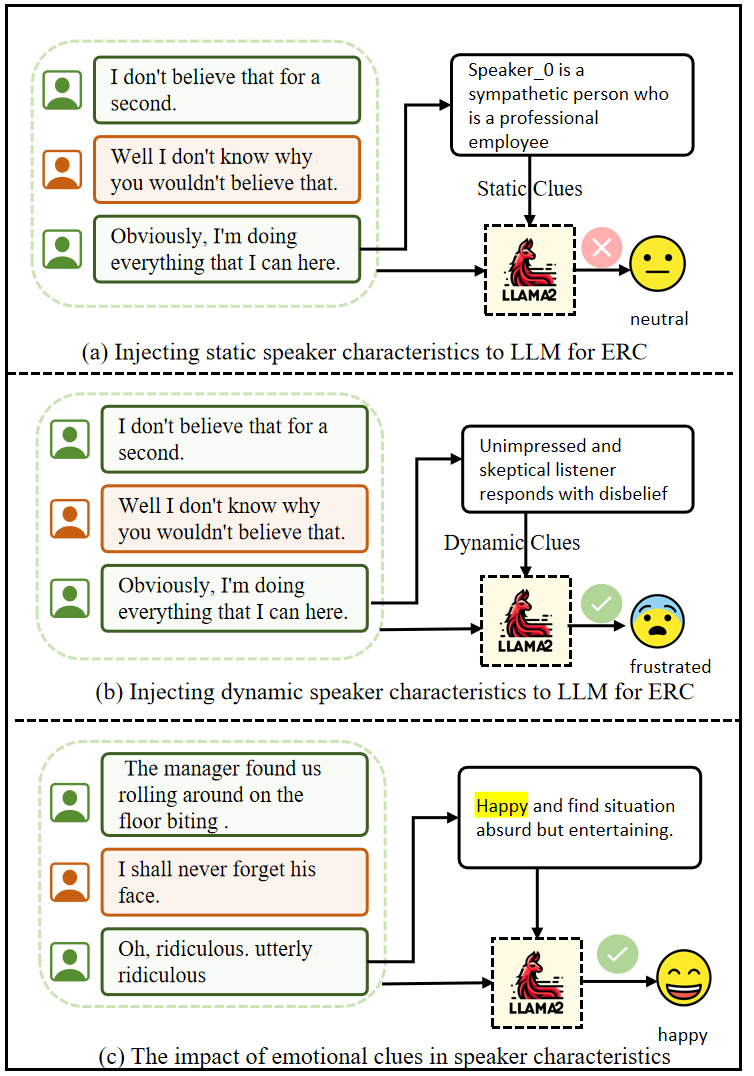}
  \caption {The case study of three samples from IEMOCAP dataset.}
  \label{fig:3}
\end{figure}

% \newpage

%fig4 case study
\section{Case Study}

In this section, we present two influence to ERC, including, speaker characteristic categories and emotional clues in speaker characteristics.

\noindent\textbf{The difference between dynamic speaker characteristics and static speaker characteristics.} 
% Figure X shows different context effection on emotion recognition in conversation. 
Figure~\ref{fig:3} (a) and (b) gives two demonstrations from IEMOCAP dataset about the same sentence affected by static speaker characteristics and dynamic speaker characteristics and then generate different emotional responses. Conversation (a) predict a neutral label due to the fact that 
speaker character is expressed as sympathy.   
In the contrast, conversation (b) generates an interactive characteristic of the current listener including some dynamic emotional clues about frustration.

% \noindent\textbf{Different emotional clues impact on ERC.}
% Figure~\ref{case}(c) shows a sentence in a conversation generates nine kinds of emotional clues through speaker relations. We can find that the results of generating emotional clues with “oReact”  or  ”xIntent”  are almost including reasonable emotion words. These emotional words indicate the emotion of the current utterance.

% \noindent\textbf{The impact of subtle discrepancies in template.}
% Figure~\ref{fig:3} (c) and (d) show the subtle differences in templates can lead to different dynamic features of the speakers. We can find that the responses of listeners generated by the template of words with `potential' will align with the emotional expression of the speaker. It indicates the `potential' represents a trend of dynamic interaction. This demonstrates the immense potential of carefully tuned templates. 

\noindent\textbf{The impact of emotional clues in speaker characteristics.}
Figure~\ref{fig:3} (c) shows the 
impact of emotional clues in speaker characteristics. We can find that the responses a word `Happy' of listeners generated  will align with the emotional expression of the speaker. It can assist the model in producing accurate results.

% We generate vocabulary related to real emotional labels to assist the model in producing accurate results

 % We can find that the responses a word `Happpy' of listeners generated  will align with the emotional expression of the speaker. It can assist the model in producing accurate results

% We can find that the results of generating emotional clues with “oReact”  or  ”xIntent”  are almost including reasonable emotion words. These emotional words indicate the emotion of the current utterance.

% \newpage
\section{Conclusion}
In this paper, we propose LaERC-S, a novel framework that explores speaker characteristics, such as mental state, behavior and persona, to promote the progress of emotion recognition in conversation (ERC). LaERC-S is well-designed with three imperative parts: speaker characteristics extraction, speaker characteristics injection and emotion analysis, all of which work in harmony to make the model reason emotional dynamics and identify emotional tendencies for each utterance in conversations. Extensive experiments on three public conversation datasets demonstrate the effectiveness and superiority of our proposed LaERC-S.

In the future work, we would like to delve into the correlation and discrepancy between speaker characteristics in form of diverse expressions. This reason is that the speaker-specific information under different perspectives presents consistent clues of an identical emotion for the utterance. These properties can make the model possess convincing explanations for emotion analysis.

% \clearpage
\section*{Limitations}

While LaERC-S has made a significant progress in adapting the LLMs for the task of emotion recognition in conversation, the current work can still be improved in the following ways. Firstly, it is important to find effective ways to maintain an efficient running cost for such large-scale embedding models. Secondly, speaker characteristics around the mental state and behavior of interlocutors have potential to be extended to other tasks in the realm of natural language processing.

\section*{Acknowledgments}
We thank the anonymous reviewers for their insightful comments. This work was supported by the Fundamental Research Funds for the Central Universities (project number: 2022FRFK060002).

% \clearpage
\bibliography{custom}

\appendix

\clearpage
\section{Appendix}

\subsection{The Details of Robustness Experiment}
\label{A.4}
In this section, we will introduce how to construct the custom dataset used in \S~\ref{RB}. 
Specifically, since the emotional labels in each original dataset are different, we need to map them to a unified label before the extracting and merging, as shown in Table~\ref{mix_label}.

\subsection{Details of the Datasets}
\label{A.1}

\noindent \textbf{IEMOCAP} is a dataset collected from improvisations or scripted scenarios, which contains 12 hours of conversation videos from 10 unique speakers. It has five sessions consisting of 151 conversations and 7,433 utterances. Each utterance is annotated with one of six emotion classes: neutral, happy, sad, excited, angry, and frustrated. 

\noindent \textbf{MELD} is another dataset including more than 13,000 video snippets from the Friends TV series. It comprises 1,433 conversations and 13,708 utterances in total. Each utterance is labeled as one of seven emotion categories: anger, disgust, fear, joy, neutral, sadness, and surprise. 

\noindent \textbf{EmoryNLP} is also based on the Friends TV series, which contains 97 episodes, 897 scenes and 12,606 utterances. Each utterances is annotated as one of seven emotion types: neutral, joyful, peaceful, powerful, scared, mad, and sad.

% \begin{table}[t]
% \renewcommand{\arraystretch}{1.0}
% \centering
% \resizebox{1.0\linewidth}{!}
% {
% \begin{tabular}{c|ccc|c}
% \hline
% Number                 & IEMOCAP            & MELD               & EmoryNLP                                & Final Emotion \\ \hline
% \multicolumn{1}{c|}{1} & happy              & joyful             & \multicolumn{1}{c|}{joyful}             & joyful        \\
% \multicolumn{1}{c|}{2} & sad                & sad                & \multicolumn{1}{c|}{sad}                & sad           \\
% \multicolumn{1}{c|}{3} & neutral            & neutral            & \multicolumn{1}{c|}{neutral}            & neutral       \\
% \multicolumn{1}{c|}{4} & angry              & angry              & \multicolumn{1}{c|}{mad}                & mad           \\
% \multicolumn{1}{c|}{5} & excited            & - & \multicolumn{1}{c|}{-} & excited       \\
% \multicolumn{1}{c|}{6} & - & surprise           & \multicolumn{1}{c|}{powerful}           & powerful      \\
% \multicolumn{1}{c|}{7} & scared             & fear               & \multicolumn{1}{c|}{frustrated}        & fear          \\
% \multicolumn{1}{c|}{8} & - & - & \multicolumn{1}{c|}{peaceful}           & peaceful      \\
% \multicolumn{1}{c|}{9} & - & disgust            & \multicolumn{1}{c|}- & disgust       \\ \hline
% \end{tabular}
% }
% \caption{Unified Label Mapping}
% \label{mix_label}
% \end{table}

\subsection{The Context Window Investigation}
\label{window}

To examine the impact of the context window in the performance, we conduct a parametric sensitivity analysis with different context window, as depicted in Table~\ref{tab:4}. We can notice that LaERC-S achieves the superior performance over InstructERC under any context window settings. This highlights the efficiency of LaERC-S on the task of ERC. For reference, in the first row of the table, LaERC-S provides a 5.01\%, 2.79\%, and 1.52\% improvements over InstructERC on IEMOCAP, MELD, and EmoryNLP, respectively. With the size increasing of the context window, the performance of both methods presents a tendency of improvement. Compared with MELD and EmoryNLP, the models on the dataset IEMOCAP present a significant improvement with the same context window. This is attributed to the average length of conversation in various datasets. The average length of IEMOCAP is longer than that of other datasets, thereby exploiting the larger window providing the necessary historical context for an improvement in performance. Although the performance discrepancy between them gradually decreases, the proposed LaERC-S still achieves significant superiority on three benchmark datasets. Therefore, we set a context window of 12 in LaERC-S to sufficiently capture the historical cotext in a conversation.

\begin{table}[t]
\renewcommand{\arraystretch}{1.0}
\centering
\resizebox{1.0\linewidth}{!}
{
\begin{tabular}{ccccc}
\hline
Number                 & IEMOCAP            & MELD               & EmoryNLP                                & Final Emotion \\ \hline
\multicolumn{1}{c|}{1} & happy              & joyful             & \multicolumn{1}{c|}{joyful}             & joyful        \\
\multicolumn{1}{c|}{2} & sad                & sad                & \multicolumn{1}{c|}{sad}                & sad           \\
\multicolumn{1}{c|}{3} & neutral            & neutral            & \multicolumn{1}{c|}{neutral}            & neutral       \\
\multicolumn{1}{c|}{4} & angry              & angry              & \multicolumn{1}{c|}{mad}                & mad           \\
\multicolumn{1}{c|}{5} & excited            & - & \multicolumn{1}{c|}{-} & excited       \\
\multicolumn{1}{c|}{6} & - & surprise           & \multicolumn{1}{c|}{powerful}           & powerful      \\
\multicolumn{1}{c|}{7} & scared             & fear               & \multicolumn{1}{c|}{frustrated}        & fear          \\
\multicolumn{1}{c|}{8} & - & - & \multicolumn{1}{c|}{peaceful}           & peaceful      \\
\multicolumn{1}{c|}{9} & - & disgust            & \multicolumn{1}{c|}- & disgust       \\ \hline
\end{tabular}
}
\caption{Unified Label Mapping}
\label{mix_label}
\end{table}

\subsection{Prompts}
\subsubsection{Definitions of key elements} 
\label{A.3.1}
We give the definition of key elements in Table~\ref{table5}.
This key elements include nine categories.

\subsubsection{Prompts for key elements}
\label{A.3.2}
% We give the definition of speaker characteristics in Table~\ref{table5}. In table 8, 
The key elements are used in template for speaker characteristics extraction and injection.
As illustrated in Table~\ref{table5} and Table~\ref{tab:accents}, we design the instruction-following templates for speaker characteristic extraction and injection, respectively. These templates provide precise descriptions for basic elements, such as ``title'', ``specific token'', ``objective'' and ``constraint'', to promote LLMs in performing the ERC task. Such a design is essential to guarantee clarity and accuracy in each stage.

\subsubsection{Details of Various Templates Design on Speaker Characteristics Extraction}
% \section{The Detail of Different Various Template Design on Speaker
% Characteristics Extraction}
\label{sec:appendix_3}
\label{A.3.3}
In the different template design shown as Table~\ref{template}, we have designed different textual expressions for each key element of speaker characteristics. For example, the key element "oReact" can be expressed as "the reaction of potential listeners", "the reaction of listeners", "the oReact of listeners ", and "the reaction of listeners to the event".
We find that we use template with "the reaction of potential listeners" word can better extract accurate speaker characteristics.

\subsection{The analysis of different emotion label' s performance}
Compared with InstructERC, our method achieves improvements in most emotion label, and presents sub-optimal performance in rare cases. (1) As for IEMOCAP, our method is superior to InstructERC across all emotion classes. The highest gain is 6.63\% on “Happy”. (2) As for remaining two datasets, our method still maintains consistent improvements, and achieves sub-optimal results on “Disgust” due to its few samples (2.6\% of the total dataset).

\label{A.2} 
\begin{table*}[t]
  \centering
  \setlength{\tabcolsep}{5.3pt}
  \begin{tabularx}{\linewidth}{c|cc|cc|cc}
  \hline
  \multirow{2}{*}{\makecell{Context\\Window}} & \multicolumn{2}{c|}{IEMOCAP} & \multicolumn{2}{c|}{MELD} & \multicolumn{2}{c}{EmoryNLP}\\
  \cline{2-7}
  & InstructERC  & LaERC-S  & InstructERC  & LaERC-S & InstructERC  & LaERC-S \\
  \hline
  1    & 56.12 &	61.13 (5.01$\uparrow$)  &	65.91 &	68.70 (2.79$\uparrow$) &	38.32 &	39.84 (1.52$\uparrow$)  \\
  5   & 68.65 &	69.97 (1.32$\uparrow$) &	66.97 &	69.21 (2.24$\uparrow$) &	40.48 &	41.96 (1.48$\uparrow$) \\
  12    & 71.39 &	72.40 (1.01$\uparrow$) &	69.15 &	69.27 (0.12$\uparrow$) &	41.37 &	42.08 (0.71$\uparrow$)  \\
  \hline
  \end{tabularx}
  \caption{Parameter analysis of the context window in the proposed method LaERC-S on three widely-used benchmark datasets. The symbol $\uparrow$ represents an improvement in performance over the compared method InstructERC.}
  \label{tab:4}
\end{table*}

\begin{table*}[h]
\centering
\begin{tabularx}{\textwidth}{ll|X}
    \hline
    \multicolumn{2}{c|}{\textbf{Key element}} & \multicolumn{1}{c}{\textbf{Description}} \\ 
    \hline
    \multirow{3}{*}{Mental-state} 
    & xIntent & The reason why the speaker would cause the event \\ \cline{2-3}
    & xReact & The reaction that the speaker would have to the event \\ \cline{2-3}
    & oReact & The reaction of listeners to the event \\ 
    \hline
    \multirow{5}{*}{Event} 
    & xWant & What the speaker may want to do after the event \\ \cline{2-3}
    & oWant & What the listener may want to do after the event \\ \cline{2-3}
    & xEffect & The effect the event would have on the speaker \\ \cline{2-3}
    & oEffect & The effect the event has on the listener \\ \cline{2-3}
    & xNeed & What the speaker might need to do before the event \\ 
    \hline
    \multirow{1}{*}{Persona} 
    & xAttr & How the speaker might be described given their part in the event \\ 
    \hline
\end{tabularx}
% \caption{Classification of Commonsense Relations into Categories}
 % Definitions of the relations in ATOMIC.
 \caption{Definitions of different key elements.}
\label{table5}
\end{table*}

\begin{table*}[h!]
\renewcommand{\arraystretch}{1.0}
\centering
\resizebox{1.0\linewidth}{!}
{
\begin{tabular}{c|c|c}
\hline
\textbf{Key element} & \textbf{Prompt}                                                                                                                                                                                                                                                                                                                                                                  & \textbf{Speaker characteristics}                                                                                 \\ \hline
xIntent           & \begin{tabular}[c]{@{}c@{}}Now You are an expert who is good at using common sense for reasoning.\\ The following conversation noted between ’\#\#\# \#\#\#’ involves several speakers.\\ \#\#\# Speaker1:"Okay, so big news." Speaker0:"What?" \#\#\#\\ Please use common sense to infer the intention of \textless Speaker0:"What?" \textgreater :\end{tabular}                       & \begin{tabular}[c]{@{}c@{}}Expecting explanation\\ or clarification\end{tabular}                     \\ \hline
xReact            & \begin{tabular}[c]{@{}c@{}}Now You are an expert who is good at using common sense for reasoning.\\ The following conversation noted between ’\#\#\# \#\#\#’ involves several speakers.\\ \#\#\# Speaker1:"Okay, so big news." Speaker0:"What?" \#\#\#\\ Please use common sense to infer the reaction of speaker in \textless Speaker0:"What?" \textgreater :\end{tabular}             & \begin{tabular}[c]{@{}c@{}}Surprised and curious\\ about the news\end{tabular}                       \\ \hline

oReact            & \begin{tabular}[c]{@{}c@{}}Now You are an expert who is good at using common sense for reasoning.\\ The following conversation noted between ’\#\#\# \#\#\#’ involves several speakers.\\ \#\#\# Speaker1:"Okay, so big news." Speaker0:"What?" \#\#\#\\ Please use common sense to infer the reaction of potential listeners in \textless Speaker0:"What?" \textgreater :\end{tabular} & \begin{tabular}[c]{@{}c@{}}Listener looks surprised \\ and excited.\end{tabular}                              \\ \hline

xEffect           & \begin{tabular}[c]{@{}c@{}}Now You are an expert who is good at using common sense for reasoning.\\ The following conversation noted between ’\#\#\# \#\#\#’ involves several speakers.\\ \#\#\# Speaker1:"Okay, so big news." Speaker0:"What?" \#\#\#\\ Please use common sense to infer the effect on speaker in \textless Speaker0:"What?" \textgreater :\end{tabular}               & \begin{tabular}[c]{@{}c@{}}Speaker 0 looks excited \\  about the news \end{tabular}                                    \\ \hline
oEffect           & \begin{tabular}[c]{@{}c@{}}Now You are an expert who is good at using common sense for reasoning.\\ The following conversation noted between ’\#\#\# \#\#\#’ involves several speakers.\\ \#\#\# Speaker1:"Okay, so big news." Speaker0:"What?" \#\#\#\\ Please use common sense to infer effect of potential listeners in \textless Speaker0:"What?" \textgreater :\end{tabular}       & \begin{tabular}[c]{@{}c@{}}Expectation arises;\\ curious minds eagerly \\ await details\end{tabular} \\ \hline
oWant             & \begin{tabular}[c]{@{}c@{}}Now You are an expert who is good at using common sense for reasoning.\\ The following conversation noted between ’\#\#\# \#\#\#’ involves several speakers.\\ \#\#\# Speaker1:"Okay, so big news." Speaker0:"What?" \#\#\#\\ Please use common sense to infer the wanted by listeners in \textless Speaker0:"What?" \textgreater :\end{tabular}             & \begin{tabular}[c]{@{}c@{}}Exciting development or \\ surprise event\end{tabular}                    \\ \hline
xAttr             & \begin{tabular}[c]{@{}c@{}}Now You are an expert who is good at using common sense for reasoning.\\ The following conversation noted between ’\#\#\# \#\#\#’ involves several speakers.\\ \#\#\# Speaker1:"Okay, so big news." Speaker0:"What?" \#\#\#\\ Please use common sense to infer the attribute of speaker in \textless Speaker0:"What?" \textgreater :\end{tabular}           & \begin{tabular}[c]{@{}c@{}} Speaker 0 is a \\ curious person \end{tabular}                                                                      \\ \hline
xwant             & \begin{tabular}[c]{@{}c@{}}Now You are an expert who is good at using common sense for reasoning.\\ The following conversation noted between ’\#\#\# \#\#\#’ involves several speakers.\\ \#\#\# Speaker1:"Okay, so big news." Speaker0:"What?" \#\#\#\\ Please use common sense to infer the wanted by speaker in \textless Speaker0:"What?" \textgreater :\end{tabular}               & \begin{tabular}[c]{@{}c@{}}Want to know\\ the big news\end{tabular}                 \\ \hline
xneed             & \begin{tabular}[c]{@{}c@{}}Now You are an expert who is good at using common sense for reasoning.\\ The following conversation noted between ’\#\#\# \#\#\#’ involves several speakers.\\ \#\#\# Speaker1:"Okay, so big news." Speaker0:"What?" \#\#\#\\ Please use common sense to infer the need of speaker in \textless Speaker0:"What?" \textgreater :\end{tabular}                 & \begin{tabular}[c]{@{}c@{}}Expecting important \\ information or reaction\end{tabular}               \\ \hline
\end{tabular}
}
\caption{Prompts of different key elements.}
\label{tab:accents}
\end{table*}

\begin{table*}[h]
\renewcommand{\arraystretch}{1.0}
\centering
\resizebox{1.0\linewidth}{!}
{
\begin{tabular}{c|c}
\toprule
\textbf{Template} & \textbf{Prompt}     

   \\ \hline
Template1         & \begin{tabular}[c]{@{}c@{}}Now You are an expert who is good at using common sense for reasoning.\\ The following conversation noted between ’\#\#\# \#\#\#’ involves several speakers.\\ \#\#\# Speaker0 : ”Hey.”Speaker1 : ”Hey.”Speaker0 : ”Esmeralda, guesswhat?”\#\#\#\\ Based on the above historical utterances, please use common sense to infer \\ \hl{the reaction of listeners to the event} in \textless Speaker1 : ”What?” \textgreater in no more than ten words of output :\end{tabular} \\ \hline
Template2         & \begin{tabular}[c]{@{}c@{}}Now You are an expert who is good at using common sense for reasoning.\\ The following conversation noted between ’\#\#\# \#\#\#’ involves several speakers.\\ \#\#\# Speaker0 : ”Hey.”Speaker1 : ”Hey.”Speaker0 : ”Esmeralda, guesswhat?”\#\#\#\\ Based on the above historical utterances, please use common sense to infer\\ \hl{the reaction of listeners} in \textless Speaker1 : ”What?” \textgreater in no more than ten words of output :\end{tabular}               \\ \hline
Template3         & \begin{tabular}[c]{@{}c@{}}Now You are an expert who is good at using common sense for reasoning.\\ The following conversation noted between ’\#\#\# \#\#\#’ involves several speakers.\\ \#\#\# Speaker0 : ”Hey.”Speaker1 : ”Hey.”Speaker0 : ”Esmeralda, guesswhat?”\#\#\#\\ Based on the above historical utterances, please use common sense to infer \\ \hl{the oReact of listeners} in \textless Speaker1 : ”What?” \textgreater in no more than ten words of output :\end{tabular}                      \\ \hline
Template4         & \begin{tabular}[c]{@{}c@{}}Now You are an expert who is good at using common sense for reasoning.\\ The following conversation noted between ’\#\#\# \#\#\#’ involves several speakers.\\ \#\#\# Speaker0 : ”Hey.”Speaker1 : ”Hey.”Speaker0 : ”Esmeralda, guesswhat?”\#\#\#\\ Based on the above historical utterances, please use common sense to infer\\ \hl{the reaction of potential listeners} in \textless Speaker1 : ”What?” \textgreater in no more than ten words of output :\end{tabular}     \\ \bottomrule
\end{tabular}
}
\caption{The samples of different templates.}
\label{template}
\end{table*}

\end{document}